\documentclass[letterpaper, 10 pt, conference]{ieeeconf}  

\IEEEoverridecommandlockouts                              

\usepackage{cite}
\usepackage{amsmath,amssymb,amsfonts}
\usepackage{algorithmic}
\usepackage{graphicx}
\usepackage{textcomp}
\usepackage{xcolor}
\usepackage{float}
\usepackage{bm}
\usepackage{mathtools}
\usepackage{multirow}
\usepackage{amsmath}
\usepackage{amssymb}
\usepackage{color}
\usepackage{nicefrac}
\usepackage{graphicx}
\usepackage{float}
\usepackage{bm}

\newcommand{\mat}[1]{\bm{#1}}
\renewcommand{\vec}[1]{\bm{#1}}

\newcommand{\greekvec}[1]{\boldsymbol{#1}}

\newcommand{\A}{\mat{A}}
\newcommand{\B}{\mat{B}}

\newcommand{\J}{\mat{J}}
\newcommand{\K}{\mat{K}}

\newcommand{\R}{\mat{R}}

\newcommand{\Q}{\mat{Q}}

\newcommand{\h}{\vec{h}}

\newcommand{\p}{\vec{p}}
\newcommand{\q}{\vec{q}}

\renewcommand{\u}{\vec{u}}

\newcommand{\x}{\vec{x}}

\newcommand{\z}{\vec{z}}

\newcommand{\ttau}{\greekvec{\tau}}

\newcommand{\Real}{\mathbb{R}}

\newcommand{\half}{\frac{1}{2}}

\newcommand{\T}{^\top}

\graphicspath{{figures/}}

\newcommand{\apex}{^\mathcal{A}}

\newcommand{\grad}{\nabla}
\newcommand{\hess}{\nabla^2}

\title{\LARGE \bf
Kinodynamic Model Predictive Control for Energy Efficient Locomotion of Legged Robots with Parallel Elasticity
}

\author{Yulun Zhuang$^{1}$, Yichen Wang$^{1}$ and Yanran Ding$^{1}$
\thanks{$^{1}$Yulun Zhuang, Yichen Wang and Yanran Ding are with the Department of Robotics, University of Michigan, Ann Arbor, MI - 48109, USA.
        {\tt\small \{yulunz, yicmwang, yanrand\}@umich.edu}}%
}

\begin{document}

\maketitle
\thispagestyle{empty}
\pagestyle{empty}

\begin{abstract}
In this paper, we introduce a kinodynamic model predictive control (MPC) framework that exploits unidirectional parallel springs (UPS) to improve the energy efficiency of dynamic legged robots. The proposed method employs a hierarchical control structure, where the solution of MPC with simplified dynamic models is used to warm-start the kinodynamic MPC, which accounts for nonlinear centroidal dynamics and kinematic constraints. The proposed approach enables energy efficient dynamic hopping on legged robots by using UPS to reduce peak motor torques and energy consumption during stance phases. Simulation results demonstrated a 38.8\% reduction in the cost of transport (CoT) for a monoped robot equipped with UPS during high-speed hopping. Additionally, preliminary hardware experiments show a 14.8\% reduction in energy consumption.
\end{abstract}


\section{Introduction}
Legged robots are poised to play an increasingly important role in applications such as search-and-rescue scenarios and planetary exploration.
The ability to navigate efficiently and dynamically in unstructured environments plays a key role in accomplishing these demanding tasks.
Some of these tasks require the robot to operate in remote areas for extended periods of time, without access to frequent battery charges.
Recent advancements in legged robot research showcase highly dynamic capabilities in quadrupeds\cite{di2018dynamic, bledt2020regularized, ding2021representation, katz2019mini, li2024cafe} and bipeds\cite{ding2022orientation, khazoom2024tailoring, dantec2022whole}. Nevertheless, the research on improving the energy efficiency of legged robot is currently under-explored.

Incorporating elastic elements in legged robots has been proven to reduce energy consumption in terms of the cost of transport (CoT)\cite{bravo2024engineering, vu2015improving, seok2013design}. Parallel elastic elements, in particular, can share external loads with motors, counteracting gravitational forces and reducing required motor torque\cite{Sreenath2011mabel, Bjelonic2023learning, liu2024dualslide}. For running robots, parallel elasticity can facilitate kinetic energy recycling by storing energy during landing, and releasing the stored energy when during push-off.
However, traditional parallel elastic components can hinder leg movement during the swing phase. 
Solutions like SPEAR \cite{Liu2015switch} and BirdBot \cite{BadriSprwitz2022bird} employ disengagement mechanisms which can alleviate this but introduce extra hardware complexity. To address this, unidirectional parallel spring (UPS) \cite{au2009powered}
offers a solution by providing assistance during the stance phase without impeding swing motion, potentially reducing energy consumption for locomotion.

Controlling running robots with parallel elasticity presents unique challenges.
While the spring-loaded inverted pendulum (SLIP) model has been widely used \cite{raibert1986legged, martin2017experimental, wensing2013high, apgar2018fast}, it struggles to incorporate explicit constraints such as motor torque saturation. 
Model predictive control (MPC) can incorporate complex system models and constraints, producing trajectories that respect physical limits. The ability to reason about state evolution within a finite horizon makes MPC suitable for controlling running robots with parallel springs, which stores and releases kinetic energy within a gait cycle.

\begin{figure}
    \centering
    \includegraphics[width=1\linewidth]{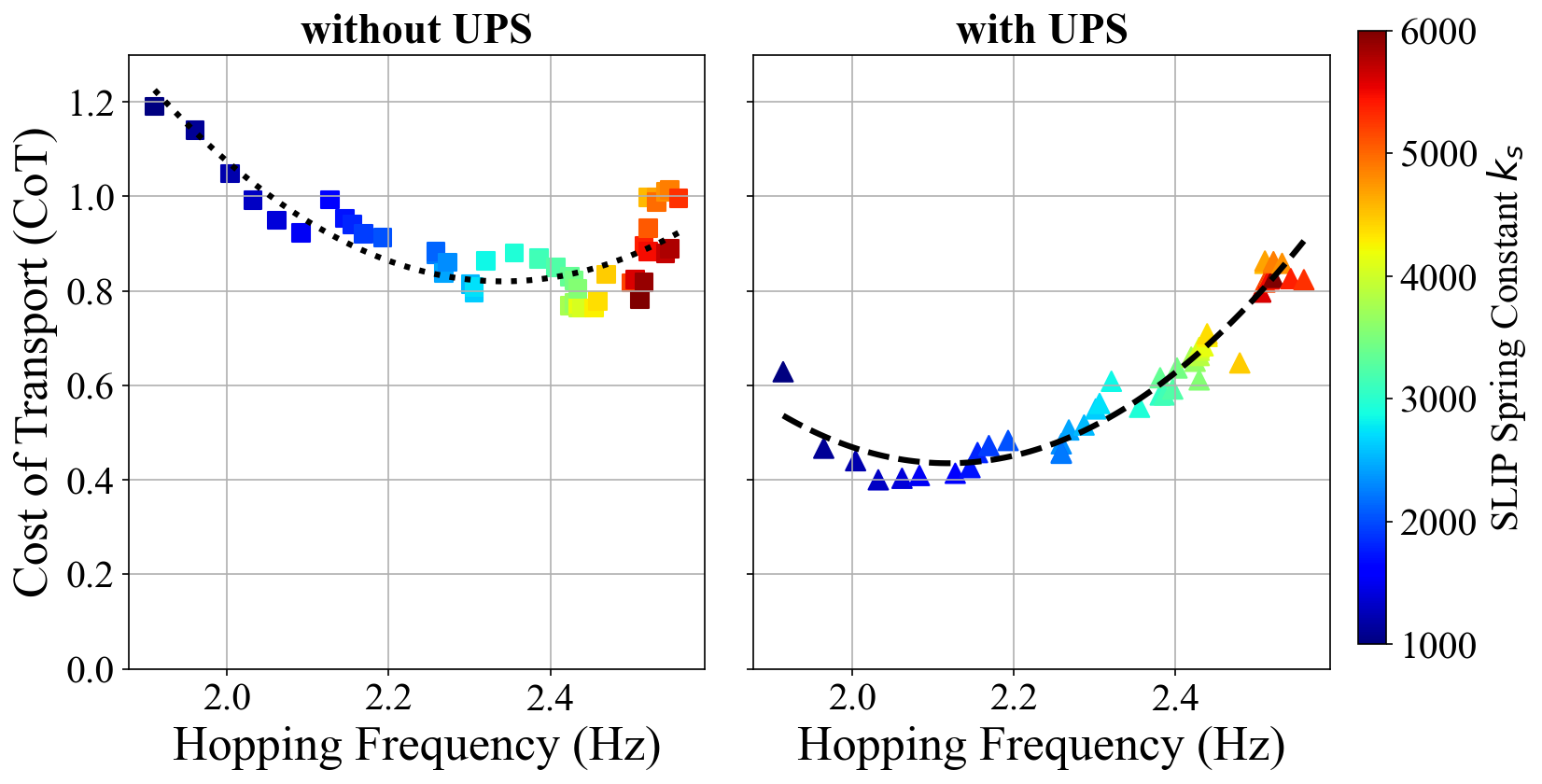}
    \caption{Simulation results showcasing the capability of the proposed kinodynamic MPC on a hopping robot with UPS. The Cost of Transport (CoT) is plotted w.r.t. hopping frequency for 10 continuous jumps at $v_x=1$ m/s. (left) CoT of the robot without UPS. (right) CoT of the robot with UPS.}
    \label{fig:cot_freq}
    \vspace{-10pt}
\end{figure}

\begin{figure*}
    \centering
    \includegraphics[width=0.9\textwidth]{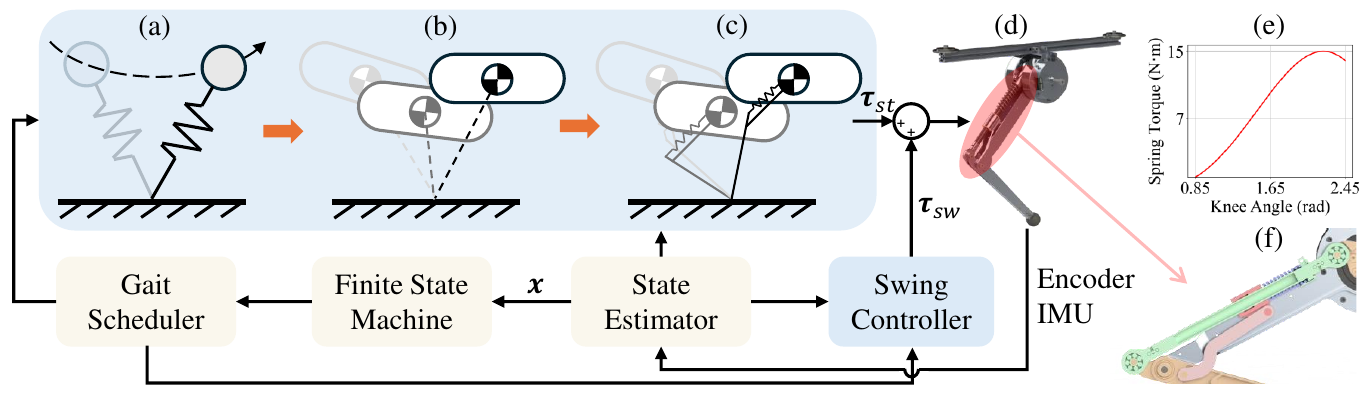}
    \caption{The system overview of the proposed hierarchical controller, where the orange arrows indicate the improvement of model fidelity. (a) The SLIP motion sketch synthesis stage generates the CoM trajectory, touchdown location and GRF. (b) The SRB MPC stage introduces rotational dynamics and friction cone constraints. (c) The kinodynamic MPC stage incorporates joint angle and joint torque to reason about UPS. (d) CAD render of the custom monoped robot. (e) The parallel spring torque as a function of knee joint angle. (f) The cross-section view of the UPS mechanism.}
    \label{fig:main}
    \vspace{-10pt}
\end{figure*}

MPC offers flexibility to accommodate various system requirements and computational resources. Convex MPC that employ simplified dynamics models, such as the single rigid body (SRB) model, are commonly used and have advantages in real-time applications because of their fast computational efficiency \cite{di2018dynamic}. Nevertheless, unmodeled aspects such as kinematic constraints cannot be accurately captured. On the other end of the spectrum, whole-body MPC \cite{mastalli2020crocoddyl, dai2014whole}, models the full dynamics and kinematics of the robot, promising maximum expressiveness and control potential, at the expense of high computational complexity.

Kinodynamic MPC \cite{dai2014whole} offers a compelling compromise between computational efficiency and control performance. Employing a simplified dynamical model allows kinodynamic MPC to inherit the fast computational speed from MPC with simple models, while having a full kinematics model allows kinodynamic MPC to produce motions that take kinematic constraints into account. However, the presence of kinematic constraints often introduces nonconvexity, necessitating solving nonlinear programming (NLP) problems. Solving NLPs in real time is challenging because they are prone to converge to local minima and infeasible solutions. To mitigate these issues, warm-starting techniques are often employed, typically by using solutions from simpler MPC problems as initial guesses.

Our work addresses the challenge of energy-efficient dynamic legged locomotion by proposing a kinodynamic MPC framework for robots with UPS.
The main contributions of this paper are: 
\begin{itemize}
    \item A hierarchical kinodynamic MPC framework that explicitly incorporates UPS dynamics, resulting in improved energy efficiency for legged locomotion.
    \item A comprehensive study of the effect of UPS on the CoT across a range of hopping speeds and frequencies, providing insights to optimal design and control strategies.
    \item Experimental results that demonstrates a 38.8\% reduction of CoT in simulation, and a 14.8\% reduction of energy consumption in preliminary  hardware experiment.
\end{itemize}

The remaining sections of this paper are organized as follows: Section~\ref{sec:method} details components for each layer of the hierarchical control framework; Section~\ref{sec:result} exhibits the results of simulation and hardware studies; Section~\ref{sec:conclusion} presents the concluding remarks and future work.

\section{Controller Design}\label{sec:method}
The goal of our controller is to improve locomotion energy efficiency by exploiting the combined dynamics of rigid linkages and compliant parallel elasticity. To achieve this, we employ a kinodynamic MPC that explicitly accounts for the dynamics of elastic springs in the joint space during the stance phase (Section~\ref{sec:kino-srb}).
To enable the kinodynamic MPC for real-time operation, a hierarchical control architecture is employed, as shown in Fig. \ref{fig:main}. First, the SLIP model (Section~\ref{sec:slip}) is used to generate a preliminary CoM trajectory, ground reaction force (GRF), and touchdown foot location. These quantities are then taken as the tracking reference by a convex MPC with the SRB model to regularize the rotational dynamics of the robot (Section~\ref{sec:srb}). The trajectories generated by the SRB MPC are then passed to the kinodynamic MPC as the initial guess for the nonlinear program, enabling a robust real-time control at 200 Hz.

\subsection{Dynamic Models}

The hierarchical control framework utilize three dynamical models as shown in Fig. \ref{fig:main}, including the SLIP (a), SRB (b), and a kinodynamic (c) model. Each model adds progressively more details, beginning with a simplified point-mass model (SLIP), extending to rotational dynamics (SRB), and culminating in full-body joint angle and torque dynamics (kinodynamic model).

The SLIP model consists of a point-mass and massless leg with a prismatic spring, and it alternates between the stance and flight phases. During the flight phase, the foot position $\bm{p}_f \in \Real^{2}$ in the world frame is steered towards the upcoming landing with a touchdown angle $\alpha$. Given CoM position $\bm{p}_c \in \Real^{2}$ and velocity $\dot{\bm{p}}_c\in\Real^{2}$, the flight phase dynamics is ballistic, while the stance dynamics is described by the ordinary differential equation
\vspace{-3pt}
\begin{equation}
    m\ddot{\bm{p}}_c =  m\bm{g} +  k_s(\|\bm{r}\|-r_0)\hat{\bm{r}},
\end{equation}
where $m$ is the mass; $k_s$ is the prismatic spring constant; $r_0$ is the rest length; $\bm{g} \in \Real^{2}$ is the gravitational vector, and $\bm{r} = \bm{p}_f - \bm{p}_c$ is the vector pointing from CoM to foot position.

Building on the CoM trajectory generated by the SLIP model, the SRB model captures the rotational dynamics of the robot torso. Since heavy actuators are placed close to the center of the torso and leg mass is less than 10\% of the total robot mass, the SRB model captures the dominant dynamical effect of the monoped robot.
The SRB dynamics is described by
\begin{equation}\label{eq:srb}
\dot{\bm{x}}=\frac{\mathrm{d}}{\mathrm{d} t}
\begin{bmatrix}
    \bm{p}_c \\
    {\theta} \\
    \dot{\bm{p}}_c \\
    \dot{{\theta}}
\end{bmatrix}
=
\begin{bmatrix}
    \dot{\bm{p}}_c \\
    \dot{{\theta}} \\
    \mathbf{f}/m+\bm{g} \\
    (\bm{r} \wedge \mathbf{f})/{I}
\end{bmatrix},
\end{equation}
where ${\theta}$ and $\dot{\theta}$ are the torso pitch angle and angular velocity, respectively; ${I}$ is the torso rotational moment of inertia; $\mathbf{f}\in\Real^2$ is the ground reaction force (GRF); $\wedge$ is the wedge product operator.

The kinodynamic model enhances the SRB model with additional quantities, including joint angle and joint torque. Further details about the kinodynamic model will be provided in the following section.

\subsection{Kinodynamic MPC for Robots with Elasticity}\label{sec:kino-srb}
Kinodynamic MPC has the unique advantage of reasoning about parallel spring torque, which is a function of joint angle, without incurring a large computational budget as in the whole-body MPC. An illustration of the kinodynamic MPC is shown in Fig. \ref{fig:main}(c).

\subsubsection{\textbf{Constraints}}
The kinodynamic MPC utilizes the nonlinear SRB dynamical model expressed in \eqref{eq:srb}, imposed as equality constraints throughout the prediction horizon $N$. The nonlinear SRB dynamics is linearized around the reference state and control trajectory, forming a linear time-varying (LTV) system with the extended state
\vspace{-2pt}
\begin{equation}
    \begin{bmatrix}
        \bm{x}_{k+1} \\ 1
    \end{bmatrix}
     = 
     \begin{bmatrix}
         \bm{A}_k & \bm{d}_k\\ \bm{0} & 1
     \end{bmatrix}
    \begin{bmatrix}
        \bm{x}_{k} \\ 1
    \end{bmatrix}
     + 
     \begin{bmatrix}
        \bm{B}_k\\ 0
     \end{bmatrix}
     \mathbf{f}_{k},
    \label{eq:linearization}\\
\end{equation}
where $\x$ is the SRB state; $\mathbf{f}$ is the GRF. The discrete dynamic matrices $\A_k\in\Real^{6\times 6}$, $\B_k\in\Real^{6\times 2}$ and $\bm{d}_k\in\Real^{6}$ characterizes the discrete linearized dynamics of \eqref{eq:srb}. The augmented state reformulates the original affine dynamics with offset $\bm{d}_k$ to linear dynamics.

To enforce kinematic feasibility and torque limits during the stance phase, the kinodynamic model incorporates joint positions $\bm{q} \in \mathbb{R}^2$ and motor torques $\bm{\tau} \in \mathbb{R}^2$ as decision variables during stance phase. Specifically, the foothold position is constrained through forward kinematics (FK):
\begin{equation}
\bm{p}_{f} = \text{FK}(\bm{p}_c, \theta, \bm{q}),
\label{eq:const_fk}
\end{equation}
ensuring that joint angles are within physically realizable configurations. 
Furthermore, the effect of parallel elasticity in the joint torque is modeled as:
\begin{equation}
\bm{\tau} + \bm{\tau}_s(\bm{q}) = \bm{S} \cdot \bm{J}^T \mathbf{f},
\label{eq:const_pea}
\end{equation}
where $\bm{\tau}_s(\bm{q})$ represents the torque contribution from the parallel elastic springs, characterized in Fig.~\ref{fig:main}(e); $\J\in\Real^{2\times 5}$ is the foot Jacobian matrix; $\bm{S}\in\Real^{2\times 5}$ is the selection matrix. 
By incorporating the parallel spring directly into the model, the MPC controller can proactively leverage the UPS. Additionally, the relationship in \eqref{eq:const_pea} arises from the massless leg assumption.
The joint angle and torque limits are incorporated as box constraints
\begin{equation}
    \q_\text{min} \leq \q \leq \q_\text{max},\ \|\ttau\| \leq \ttau_\text{max}\label{eq:tau_limit},
\end{equation}

To prevent foot slippage, the GRF has to respect the following friction cone constraints
\begin{align}
    &-\mu\ \mathrm{f}_{k, z} \leq \mathrm{f}_{k, x} \leq \mu\ \mathrm{f}_{k, z}\label{eq:const_friction}\quad \quad\\
    & 0\leq \mathrm{f}_{k,z} \leq c_k\ \mathrm{f}_\text{max},\ \forall k,\quad \quad
    \label{eq:const_contact}
\end{align}
where $\mu$ is the coefficient of friction; $c_k$ indicates the contact schedule, such that $c_k=1$ indicates contact at step $k$ and $c_k=0$ indicates no contact; $\mathrm{f}_\text{max}$ is the maximum vertical component of the generated GRF.

\subsubsection{\textbf{Objectives}}
The objective function is defined as the weighted sum of state and GRF tracking errors, and the sum of squared joint torques
\begin{equation}
    \ell_{KD}(\x, \u) = \|\x-\x^\text{des}\|_{\Q}^2 + \|\mathbf{f} - \mathbf{f}^\text{des}\|_{\R_{\mathbf{f}}}^2 + \|\ttau\|^2_{\R_{\ttau}},
    \label{eq:obj_kino_srb}
\end{equation}
where
$\|\cdot\|_{\bm{P}}^2$ is the ${l}^2$ norm weighted by a positive-definite diagonal matrix $\bm{P}$; the control inputs $\u = [\mathbf{f}\T, \ttau\T]\T$ are a concatenated vector of GRF and motor torques; superscript $(\cdot)^\text{des}$ indicates desired values.

\subsubsection{\textbf{Kinodynamic MPC Formulation}}
The objective function and constraints are employed to formulate the kinodynamic MPC, which is transcribed as an NLP
\begin{align}
    \min_{\x_k, \u_k, \q_k}~~ &\gamma^N \ell_{KD}^f(\x_N) + \sum_{k=0}^{N-1} \gamma^k \ell_{KD}(\x_k, \u_k)\label{eq:kino_srb}\\
    \text{s.t.}~~~~~
    &\text{linearized discrete dynamics:~\eqref{eq:linearization}}\nonumber\\
    &\text{nonlinear kinematics constraint:~\eqref{eq:const_fk}}\nonumber \\
    &\text{joint torque constraint with UPS:~\eqref{eq:const_pea}}\nonumber\\
    &\text{joint position and torque limits:~\eqref{eq:tau_limit}}\nonumber\\
    &\text{friction cone constraints:~\eqref{eq:const_friction},\ \eqref{eq:const_contact}}\nonumber\\
    &\x_0 = \x_{\text{initial}}\nonumber,
\end{align}
where $N$ is the prediction horizon; The decay rate $\gamma \in $ (0, 1) is a parameter that addresses the model inaccuracy issue further down the prediction horizon; $\ell^f$ is the terminal objective function that is a function of terminal state $\x_N$.

\subsection{SLIP for Motion Sketch Generation}\label{sec:slip}

While kinodynamic MPC can reason about the dynamical effect of the UPS, it presents challenges in solving the transcribed NLP online. To address this issue, we employ a hierarchical architecture to provide initial guesses to warm-start the NLP.

The SLIP model is used as the template to generate a motion sketch. Specifically, given a desired forward velocity $v_{c,x}$, the CoM state trajectory, ground reaction force, and the next touchdown location parameterized by the touchdown angle $\alpha$, are generated by the SLIP model. The apex state $\apex\x = [p_{c, z}, v_{c, x}] \in\Real^2$ is defined as the state when the robot reaches its apex during the flight phase, where $p_{c, z}$ is the apex height and $v_{c, x}$ is the apex velocity. Given the touchdown angle $\alpha$, the apex-to-apex return map is defined as $\apex\x_{n+1} = \h(\apex\x_n, \alpha_n)$, which maps the current apex state to the next apex state. 

To achieve periodic running at the desired speed, we find the optimal state-control pair ($\apex\x^*, \alpha^*$) that minimize the difference between current apex state and the next apex state
\begin{align}
    \min_{p_{c,z}, \alpha}\quad \|\apex\x_{n} - \h(\apex\x_{n}, \alpha_{n})\|^2,
    \label{eq:slip_nonlstsq}
\end{align}
where $p_{c,z}, \alpha$ are the decision variables, $\apex\x^+$ is the next apex state.
This problem can be formulated as a nonlinear least-squares problem \cite{wensing2013high} and solved efficiently.

We constructed a gait library offline for an array of desired apex velocities, where we computed the optimal ($\apex\x^*, \alpha^*$) pair and the corresponding deadbeat feedback controllers \cite{wensing2013high}. The deadbeat controller is obtained by locally linearizing the return map with the linear feedback control law is
\begin{equation}
    \alpha = \alpha^* + \K({\apex}\x - {\apex}\x^*),
    \label{eq:slip_feedback_law}
\end{equation}
where ${\apex}\x$ is the measured apex state and $\K\in\Real^{1\times 2}$ is the Jacbobian matrix computed using finite difference. Given an arbitrary desired apex velocity, the gait library will query the optimal state-control-gain tuple through interpolation in real-time. This information will be used to simulate the SLIP model forward to obtain the CoM state, ground reaction force, and touchdown location for the subsequent SRB MPC.

\subsection{Convex SRB MPC for Fast Warm-Start}\label{sec:srb}

Given the motion sketch from SLIP model as the reference, a convex MPC based only on the SRB model, in Fig.~\ref{fig:main}(b), is formulated.
The nonliear term $\bm{r}\wedge \bm{\mathrm{f}}$ in the SRB dynamics in~\eqref{eq:srb} motivates the following convex approximations.
Since there is no reference of rotational dynamics for direct linearization,
we evaluate $\bm{r}$ at desired state $\bm{p}_c^{\text{des}}$ and foothold locations $\bm{p}_f^{\text{des}}$, which are computed ahead of time.
In this way the SRB dynamics becomes linear on states and controls and can be discretized as the same form in \eqref{eq:linearization}.

To minimize the tracking errors between reference states and controls, the objective function is defined as
\begin{equation}
    \ell_{SRB}(\x, \mathbf{f}) = \|\x-\x^\text{des}\|_{\Q}^2 + \|\mathbf{f} - \mathbf{f}^\text{des}\|_{\R_{\mathbf{f}}}^2,
    \label{eq:obj_srb}
\end{equation}

By employing only the relaxed SRB dynamics and friction cone constraints, the convex MPC is formulated as 
\begin{align}
    \min_{\x_k, \mathbf{f}_k}~~ \gamma^N &\ell_{SRB}^f(\x_N) + \sum_{k=1}^{N-1} \gamma^k \ell_{SRB}(\x_k, \mathbf{f}_k) \label{eq:srb_mpc}\\
    \text{s.t.}~~~~~~~
    &\text{linearized discrete dynamics:~\eqref{eq:linearization}}\nonumber\\
    &\text{friction cone constraints:~\eqref{eq:const_friction},\ \eqref{eq:const_contact}}\nonumber\\
    &\x_0=\x_{\text{initial}}\nonumber,
\end{align}
which can be transcribed as a quadratic program (QP) that can be solved efficiently using off-the-shelf solvers.

\subsection{Energy Consumption Modeling and Metric}
This work uses the CoT $\coloneqq P/(mgv)$\cite{seok2013design} as a metric for the energy efficiency of locomotion. CoT is the ratio of the the average power to the product of weight and forward velocity. The energy consumption model assumes that the total power is comprised of mechanical power and Joule heating; the former can be calculated as the product of joint torque and joint angular velocity. The thermal loss rate is modeled by $i_m^2R$, where $R=0.17~\Omega$ is the equivalent winding resistance. The motor torque $\tau_m = K_t i_m$ is the product of torque constant $K_t$ and motor current. We assume that the battery can not be recharged by back-EMF, so only the positive power $P^{+}$ is accumulated\cite{ackerman2013energy}. 
Therefore, the power consumption is modeled as 
\begin{equation}
    P^+_k = \text{max}(\ttau_k \T \dot{\q_k}, 0) + \|\ttau_k/K_t\|^2 R,
    \label{eq:power}
\end{equation}
where $\text{max}(\cdot) $ ensures the nonnegativity of motor power. $P^+$ is averaged over a finite period to calculate the CoT; $P^+$ is integrated over a duration to calculate the total energy.

\subsection{Solution Strategy of the Optimization Problem}\label{sec:sqp}
This section focuses on solving the kinodynamic MPC. SLIP motion can be obtained by solving a nonlinear least-square problem, and SRB MPC is transcribed as a convex optimization, both of which can be solved in real-time. In contrast, solving the kinodynamic MPC is more challenging since it involves solving NLP online. 

To provide robust solutions in real-time, we utilize the sequential quadratic program (SQP) method. A general NLP can be described with decision variables $\z$, the objective function $J(\z)$, and inequality constraints $\bm{G}(\z)$. It can be locally approximated as a QP subproblem
\begin{subequations}
\begin{align}
    \min_{\bm{d}} ~~~
    & J(\z) + \grad J(\z)\T\bm{d} + \half \bm{d}\T\hess \mathcal{L}(\z, \bm \lambda)\bm{d}\\
    \text{s.t.} ~~~\,
    & \bm{G}(\z) + \grad\bm{G}(\z)\T\bm{d} \geq \bm{0},
\end{align}
\label{eq:sqp}%
\end{subequations}
where the Lagrangian function of this problem is $\mathcal{L}(\bm{z}, \bm{\lambda}) = J(\bm{z}) - \bm\lambda\T \bm{G}(\bm{z})$;
$\bm{\lambda}$ is the Lagrangian multiplier;
$\bm{d}$ is the step direction; 
$\grad J(\cdot)$ is the gradient of the objective function;
$\hess \mathcal{L}(\cdot)$ is the Hessian of the Lagrangian;
$\grad \bm{G}(\cdot)$ is the Jacobian of constraint with respect to $\z$.

We solve~\eqref{eq:sqp} iteratively using the OSQP\cite{stellato2020osqp} solver, which provides low-accuracy solutions within a few computationally cheap iterations with a warm-started initial guess. Solving SQP problems using the OSQP solver allows us to trade the solution accuracy for accelerated computational time by solving a fixed number of iterations at high rates. Overall, the SQP approach allows us to enforce leg kinematics and parallel elasticity constraints in real-time with sufficient accuracy. 

\subsection{Swing Leg Control}\label{sec:swing}

The swing controller tracks the desired foot trajectory in the world frame. The desired swing foot trajectory is computed using the desired touchdown angle $\alpha$ given by the deadbeat policy \eqref{eq:slip_feedback_law}. A Bézier polynomial is used to interpolate the liftoff and touchdown foot positions with ground clearance in the world frame to obtain the desired foot position $\p_f^\text{des}$. The desired joint angle $\q^{des}$ is calculated using inverse kinematics (IK) for the proportional-derivative (PD) control scheme
\begin{equation}
    \ttau = \K_P(\q^\text{des} - \q) - \K_D  \dot\q,
\end{equation}
where $\K_P, \K_D\in\Real^{3\times 3}$ are diagonal positive definite proportional and derivative gain matrices; $\q$ and $\dot\q$ are the measured joint state from encoder. The reference joint velocity is set to zero to ensure stability. In addition, a centroidal momentum based algorithm \cite{orin2008centroidal} is used to ensure that the dynamical effect of a swing leg does not affect apex detection. When a contact event is detected, the finite state machine (FSM) switches to the stance controller, and the SLIP reference is reset.

\subsection{Controller Details}
The desired values $\x^\text{des}_k$ and $\mathbf{f}^\text{des}_k$ in \eqref{eq:obj_kino_srb} are composited of CoM states and GRFs reference interpolated from the SLIP model, while the reference rotational states are set to zero. 
Due to the discrete changes between stance and flight phases, the prediction timestep should consider these discrete modes. The timestep $\Delta t_k$ is adaptively segmented based on the algorithm in \cite{bledt2020regularized}, by taking a nominal discretization timestep and the predefined contact schedule to calculate the closest set of timestep segmentation while respecting the exact contact schedule.

\section{Results}\label{sec:result}

\begin{figure}[htb]
    \centering
    \includegraphics[width=.8\linewidth]{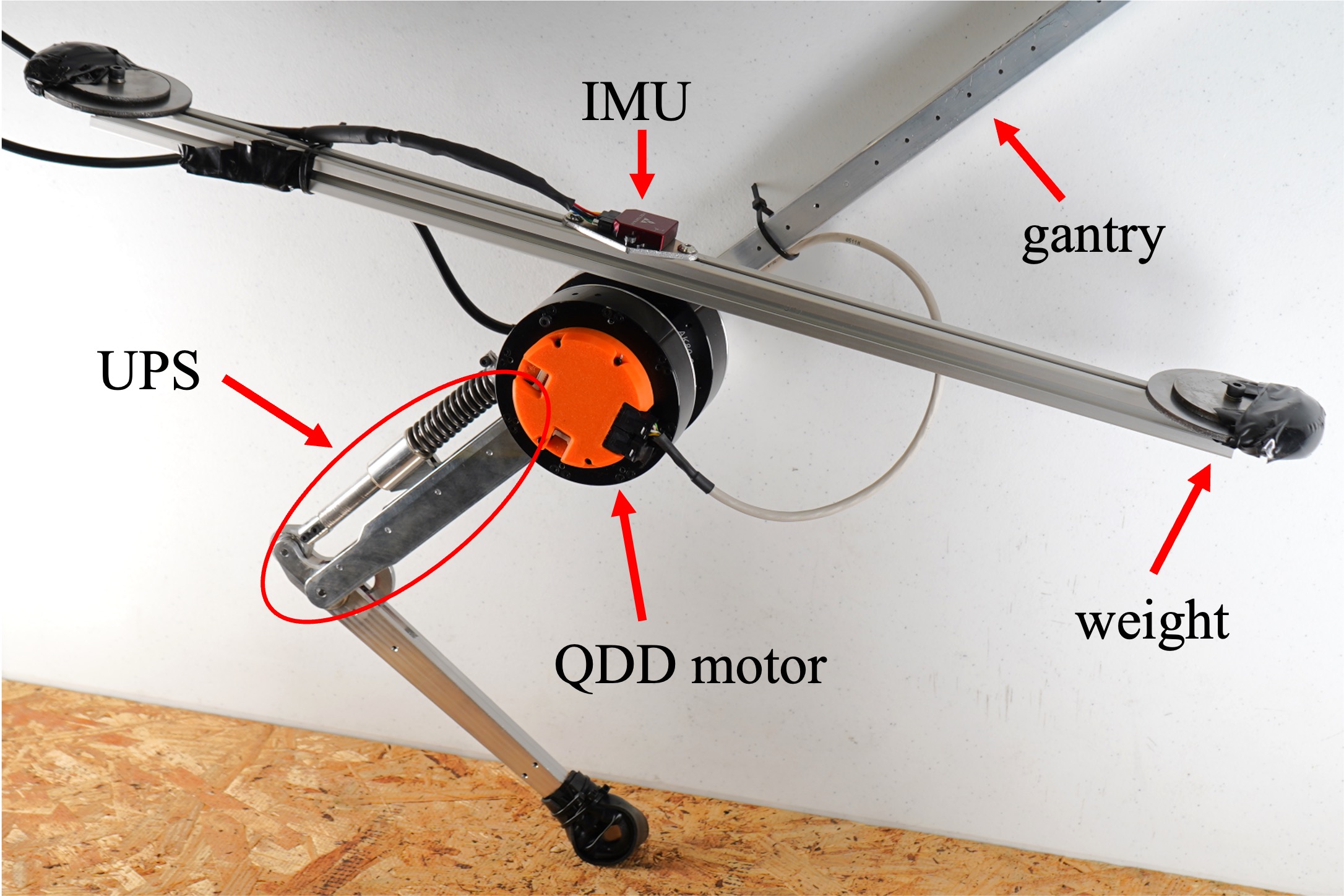}
    \caption{Monoped with Unidirectional Parallel Spring (MUPS) is a custom robot  platform to validate our proposed kinodynamics MPC framework.}
    \label{fig:robot}
    \vspace{-10pt}
\end{figure}

\subsection{The MUPS Robot}
The Monoped with Unidirectional Parallel Spring (MUPS) is a hopping robot, which is our prototype robot for the study of the design and control of legged robots with UPS. As shown in Fig. \ref{fig:robot}, MUPS comprises a hip and a knee motor and a torso connected to a gantry to constrain its motion to be within the sagittal plane. Compared with other monopeds with a torso fixed to the gantry \cite{ding2017design}, MUPS has a free-rotating torso that connects to the gantry through bearings. The hip and knee motors are quasi-direct drive (QDD) motors produced by T-Motor, which provides high torque density and control bandwidth \cite{katz2019mini}. The knee motor is located coaxially with the hip motor, resulting in low leg mass and inertia. The knee motor torque is transmitted to the knee joint via a four-bar linkage, which is designed to incorporate the UPS mechanism, whose elastic component is a linear compression spring. The UPS mechanism uses an additional link and slider to compress a spring fitted over the coupler, storing energy. This configuration saves weight by using the coupler link to both transmit power from the motor and to fit the spring and slider.

\subsection{Computation Setup}

\begin{table}[htb]
\caption{Constants of the robot and parameters of the controller}
\label{tab:param}
\centering
\begin{tabular}{ccc|cc}
\hline
\textbf{Const.} & \textbf{Value} & \textbf{Unit} & \textbf{Param.} & \textbf{Value}  \\ 
\hline
$k_s$ & 1500 & N$/$m    & $\mu$     & 0.7   \\
$r_0$ & 0.32 & m        & $\K_P$    & [40, 40]   \\
$m$   & 2.5  & kg       & $\K_D$    & [0.5, 0.5]  \\
$I$   & 0.05 & kg$\cdot$m$^2$ & $\Q_{\x}$   & [10, 10, 1]   \\
$\q_\text{min}$     & [0, $-$2.45]  & rad & $\Q_{\dot\x}$ & [1, 0, 0.1]  \\
$\q_\text{max}$     & [$\pi$/2, $-$0.85] & rad & $\R_{\mathbf{f}}$ & [10$^{-\text{5}}$, 10$^{-\text{5}}$]  \\
$\ttau_\text{max}$ & [25, 25] & N$\cdot$m & $\R_{\ttau}$   & [10$^{-\text{5}}$, 10$^{-\text{5}}$]   \\ 
\hline
\end{tabular}
\end{table}

The MPC formulations are implemented in CasADi\cite{andersson2019casadi} and complied via C code generation to speed up for real-time performance. The controller is running on a desktop computer with a 13th Gen Intel Core i7-13700 CPU and communicated to the robot via LCM\cite{huang2010lcm} at 200~Hz.
The SLIP-based gait library for a speed range from -3 m/s to 3 m/s with an interval of 0.1 m/s can be generated offline within 2 seconds.
The physical parameters of the robot and controller parameters are summarized in Table~\ref{tab:param}.

\subsection{Simulation Experiments}
The proposed controller is verified in the MuJoCo \cite{todorov2012mujoco} physics simulator, where the ground truth state is used for the controller. LCM communication pipeline is used to read robot states from MuJoCo and send joint torque commands from MATLAB.

\begin{figure}[htb]
    \centering
    \includegraphics[width=0.9\columnwidth]{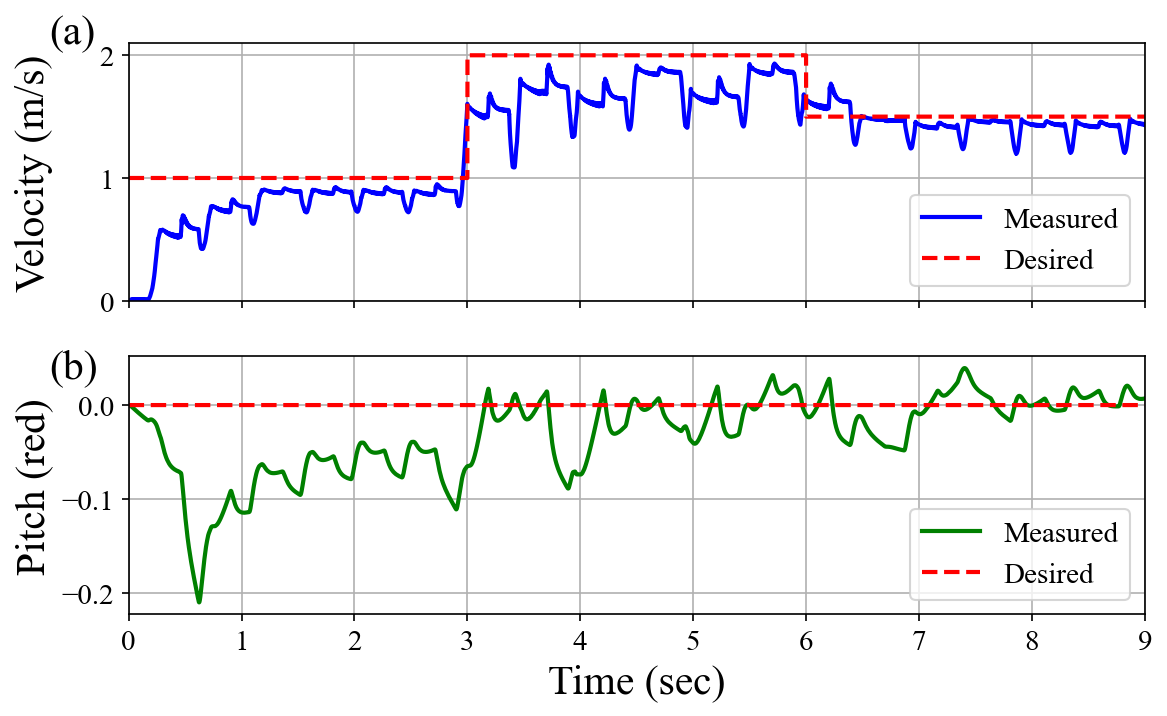}
    \caption{Simulation results for velocity tracking. The robot starts from $v_x=0$ m/s and hops to $v_x^\text{des}=1.0, 2.0, 1.5$ m/s. (a) Desired and measured forward velocity; (b) Desired and measured torso pitch angle.}
    \label{fig:vel_tracking}
\end{figure}

\subsubsection{\textbf{Velocity Tracking}}
A velocity tracking experiment is conducted to demonstrate that the robot can follow closely to the commanded velocity.
As it is shown in Fig.~\ref{fig:vel_tracking}, the robot starts from a zero velocity to track a desired velocity profile as shown in Fig.~\ref{fig:vel_tracking}(a), where the MUPS converged to the desired velocity within 3 jumps. Fig.~\ref{fig:vel_tracking}(b) shows that the torso pitch angle was regularized when encountering large velocity target changes. The maximum pitch deviation of 0.21 rad happened at 0.7 s due to fast leg retraction when the robot accelerated from stationary to 1 m/s.

\begin{figure}[htb]
    \centering
    \includegraphics[width=.8\linewidth]{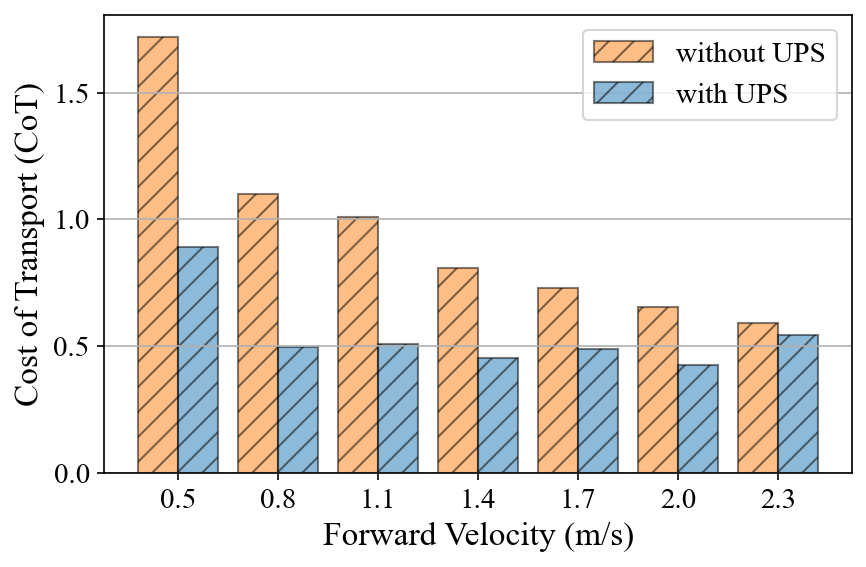}
    \caption{Simulation study that investigates the effect of UPS on CoT with increasing commanded forward velocity. CoT is the average value for 10 consecutive jumps.}
    \label{fig:cot_vx}
\end{figure}

\subsubsection{\textbf{CoT v.s. Forward Velocity}}
A simulation study is conducted to investigate the effect of UPS on CoT w.r.t. running speed, where the robot is commanded to run forward with a constant speed range from 0.5 m/s to 2.3 m/s. The data is only recorded when the robot reached the commanded speed.
As shown in Fig.~\ref{fig:cot_vx}, with increasing forward velocity, MUPS achieved low CoT across the entire velocity range, while monoped without UPS assistance resulted in higher CoT. On average, engaging UPS results in a CoT reduction of 38.8\%.

\subsubsection{\textbf{CoT v.s. Hopping Frequency}}
A simulation study is conducted to investigate the effect of UPS on CoT w.r.t. hopping frequency, where the robot is commanded to run at 1 m/s for 10 continuous jumps. As shown in Fig.~\ref{fig:cot_freq}, the hopping frequency increases from 1.9 Hz to 2.7 Hz by varying the SLIP spring constant from 1000 N/m to 6000 N/m. Lower CoT is realized by engaging UPS across the board. Without UPS, the monoped reached a minimum CoT of 0.79 the hopping frequency of 2.3 Hz. In contrast, the MUPS achieved a minimum CoT of 0.39 at a frequency of 2.1 Hz. 
With UPS engaged, a more compliant spring constant is required, which lightens the load on the motor.

\subsection{Hardware Experiment}

\subsubsection{\textbf{System Setup}}
The MPC controller runs real-time on an upper desktop computer, while a Raspberry Pi relays torque commands to the QDD motors and receives sensor readings from motor encoders and IMU.
The robot state is estimated via a Kalman filter which also runs on the Pi at 1~kHz. 
The robot's motion is constrained in the sagittal plane using a gantry system while allowing unconstrained X- and Z- direction movement, and torso rotation.
The robot hardware setup is shown in Fig~\ref{fig:robot}. 

\begin{table}[htb]
\caption{Hardware results of energy consumption comparison}
\centering
\label{tab:real_energy}
\def\arraystretch{1.5}
\begin{tabular}{cccc}
\hline
\multirow{2}{*}{} & \multicolumn{2}{c}{\textbf{Consumed Energy} (J)} & \multirow{2}{*}{\textbf{Reduction}} \\ \cline{2-3}
    & w/ UPS & w/o UPS &         \\ \hline
\textbf{Average} & 81.19 $\pm$ 1.01 & 95.32 $\pm$ 3.43 & 14.8\% \\
\hline
\end{tabular}
\end{table}

\subsubsection{\textbf{Preliminary Hopping Results}}
As an initial investigation, we evaluate the efficacy of the UPS on energy efficiency through an in-place hopping experiment. 
Since there is no horizontal displacement, energy consumption is used as the primary metric for efficiency comparison instead of CoT. The robot starts from a standing position and performs 10 consecutive hops while maintaining torso stability. Each condition, with and without the UPS, is tested three times. The total energy consumed in each trial is computed using \eqref{eq:power}, with results summarized in Table~\ref{tab:real_energy}. On average, integrating the UPS reduces energy consumption by 14.8\%, demonstrating its effectiveness in improving energy efficiency.

\begin{figure}[htb]
    \centering
    \includegraphics[width=.8\linewidth]{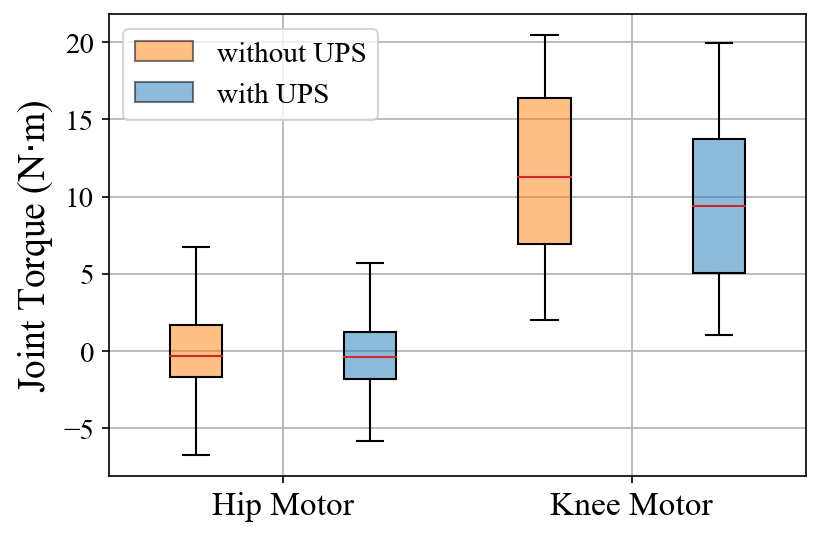}
    \caption{Hardware results of torque distributions of hip and knee motors for 10 continuous in-place jumps with and without the UPS enabled.}
    \label{fig:real_tau_distrib}
\end{figure}

Fig.~\ref{fig:real_tau_distrib} presents the measured torque distributions for the hip and knee motors. The results show that the UPS shifts the knee motor torque distribution toward zero, reducing both the average and peak torques. Specifically, the average knee torque decreases from 11.4 Nm to 9.4 Nm, while the upper quartile torque is reduced from 16.4 Nm to 13.7 Nm. This reduction indicates that the parallel spring absorbs and redistributes part of the load, alleviating the burden on the actuators. The hip torque remains largely unaffected, suggesting that the UPS primarily influences the knee joint dynamics

\section{Conclusion and Future Works}\label{sec:conclusion}

In conclusion, this work demonstrates the effectiveness of leveraging kinodynamic MPC to improve the energy efficiency and agility of a monoped robot equipped with UPS. The proposed hierarchical control framework successfully integrates joint elasticity into the optimization process, enabling real-time agile running performance. Within the hierarchy,  SLIP-based trajectory as the CoM reference and convex SRB MPC as initial guesses are combined to warm start the kinodynamic MPC. Through simulations and hardware experiments, the UPS-assisted monoped robot achieved an average 14.8\% improvement in energy efficiency, measured by the CoT. Furthermore, the UPS mechanism's ability to reduce motor torques while maintaining robust performance across varying speeds and frequencies highlights its potential for scaling the approach to more complex legged robots such as biped or humanoid robots, offering practical energy savings for industrial and exploratory applications.

The discrepancy of energy reduction between simulation and real-world experiments likely arises from unmodeled real-world factors such as joint friction, electrical losses in the actuators, and deviations from ideal spring behavior. Future work could focus on refining the system model to better capture these non-idealities, thereby improving the accuracy of energy efficiency predictions and control performance.
Additionally, further optimization of the UPS design—such as tuning its stiffness properties or introducing adaptive engagement mechanisms—could enhance its effectiveness across a broader range of speeds and loading conditions.

\section*{Acknowledgment}
We would like to thank Peter Redman, Yue Qin and Jared Grinberg for their help in developing part of the software and hardware infrastructure that enabled the experiments in this paper. This project is partially funded by the Department of Robotics, University of Michigan, and partially supported by National Science Foundation, under grant 2427036.

\addtolength{\textheight}{-0cm}   



\bibliographystyle{IEEEtran}
\bibliography{reference}

\end{document}